\documentclass{article}
\usepackage[utf8]{inputenc}
\usepackage{graphicx}
\usepackage{amsmath}
\usepackage{amssymb}

\usepackage{tikz}
\usepackage{calc}
\usetikzlibrary{calc,arrows.meta}
\usepackage{float}
\usepackage{rotating}
\usepackage{pgfplots}

\usepackage[comma,authoryear]{natbib}
\usepackage{hyperref}

\usepackage{graphicx}
\usepackage{amsmath}
\usepackage{amssymb}
\usepackage{amsthm}
\usepackage{bm} 
\theoremstyle{definition}

\usepackage{paralist} 
\usepackage[none]{hyphenat} 

\usepackage{tabularx,makecell} 

\usepackage{booktabs} 

\usepackage{tikz}
\usepackage{calc}
\usetikzlibrary{calc,arrows.meta}
\usepackage{float}
\usepackage{rotating}
\usepackage{pgfplots}
\tikzset{
  >={To[length=6pt]}
  }

\usepackage{minitoc}

\usepackage{pdfpages}

\begin{document}

\title{Abstaining Machine Learning -- \\
Philosophical Considerations}
\author{
    Daniela Schuster \\
    University of Konstanz \\
    Konstanz, Germany \\
    \texttt{daniela.schuster@uni.kn}
}
\date{2024}
\maketitle

\begin{abstract}
This paper establishes a connection between the fields of machine learning (ML) and philosophy concerning the phenomenon of behaving neutrally.
It investigates a specific class of ML systems capable of delivering a neutral response to a given task, referred to as abstaining machine learning systems, that has not yet been studied from a philosophical perspective.
The paper introduces and explains various abstaining machine learning systems, and categorizes them into distinct types.
An examination is conducted on how abstention in the different machine learning system types aligns with the epistemological counterpart of suspended judgment, addressing both the nature of suspension and its normative profile. Additionally, a philosophical analysis is suggested on the autonomy and explainability of the abstaining response.
It is argued, specifically, that one of the distinguished types of abstaining systems is preferable as it aligns more closely with our criteria for suspended judgment. Moreover, it is better equipped to autonomously generate abstaining outputs and offer explanations for abstaining outputs when compared to the other type.
  \end{abstract}

\section{Introduction}\label{AML-Introduction}
This paper investigates neutral behavior in machine learning (ML). 
In particular, we investigate so-called \emph{Abstaining Machine Learning} (AML) systems \citep{Campagner.2019}, sometimes also referred to as \textit{ML with a reject option} \citep{hendrickx.2021}, and draw parallels to the philosophical use of suspension of judgment. While in philosophy, we mostly employ the term ``suspension,'' in the context of machine learning, we will refer to the neutral behavior with the term ``abstention'' following the standard terminology within this field.\\
\indent
To fruitfully bridge the phenomena in these fields, it is beneficial to view both as neutral behaviors towards certain questions that is currently ``under discussion.''\footnote{As \citet{ferrari.2022} adopts the term ``question under discussion'' from \citet{roberts.1996}, it is predominantly used in contexts involving multiple interlocutors who align on a common goal by accepting a question as under discussion. Our considerations are limited to one single subject. Still, we employ the term ``question under discussion'' (or QUD) to \emph{fix} a specific question we wish to be seen as the object of epistemic consideration for the moment, occasionally also to differentiate it from other potential questions within the context.}
We consider questions like: ``Which dog breed is displayed in this image'', ``Is this tumor malignant or benign?'' or ``Is this person creditworthy?'', which have a finite set of well-defined, full answers $A$. This set consists of all the \textit{defined} possible answers to the question. For $Q_1=$ ``Is this tumor malignant or benign?'', $A_1=\{\textit{malignant, benign}\}$. For the question $Q_2=$ ``Which dog breed is displayed in the image?'’, possibly $A_2=\{\textit{Husky, Labrador, Dachshund, Retriver}\}$.
And for propositional questions like ``Is this person creditworthy?'' the set can simply be $\{\textit{yes, no}\}$.
In the context of Machine Learning, the answers are typically identified with \textit{outputs}. To indicate the use of a term as an output, we will employ a typewriter font, i.e.,\  \texttt{malignant} and \texttt{Labrador}, and so on.
\\
\indent
In this work, we focus on those situations in which \textit{none of the answers} from the answer set is selected. Instead, the question is addressed with a response that expresses neutrality, uncertainty, or indecision about the correct answer.\\
\newline
In philosophy, this neutrality is commonly described with the term ``suspension of judgment,'' which is usually characterized as a doxastic, mental stance whose counterparts are belief and disbelief. While belief and disbelief express those doxastic positions that are accompanied by some certainty or decisiveness about a question $Q$ and its correct answer, suspension expresses neutrality and indecision about $Q$.\footnote{The analogy between belief and disbelief can be drawn best for propositional questions that have only ``Yes'' and ``No'' in their answer set.}\\
\indent
In machine learning, neutral outputs are described with the term ``abstention.''
Traditionally, for an ML algorithm tasked with answering a question $Q$ of the above type, the set of possible outputs is equal to the set of the defined answers $A$.\footnote{At least this is so for a classification problem, which we will concentrate on.} For the question about the dog breed, the algorithm could output \texttt{Husky}, and for the question about the tumor, the algorithm could output \texttt{benign}. Abstaining machine learning algorithms are additionally able to output an \texttt{abstention} response, which is not a member of the defined answers in $A$.\\
\newline
\noindent
In bringing the two fields and respective debates together, this paper starts to fill a gap in the philosophy of AI literature. Philosophy of AI is concerned with describing and evaluating AI systems with the help of philosophical terms, norms, and debates. So far, this has not been done for abstaining machine learning, although this area provides an enormous potential for philosophical investigations.\\
\indent
Abstaining ML is a field in ML research that is still considered only by a relatively small group of researchers \citep{Campagner.2019, Ferri.2004} and largely unknown to philosophers. This is surprising, considering that AML systems show a promising way to uncover and deal with uncertainties in decision processes.
As argued by
\citet{phillips.2020}, the awareness of its own knowledge limits is one key principle of an explainable artificial intelligence. Abstaining Machine Learning provides a direct method for explicitly defining these knowledge limits and communicating them to users.
\\
\indent
In this paper, we intend to enhance the awareness and comprehension of abstaining machine learning among both AI researchers and philosophers. By doing so, we aim to contribute to the fields of trust and explainability in AI systems by underlining the significance of uncovering and effectively communicating uncertainties and the limits of knowledge.
\\
\newline
\noindent
The way in which the paper aims to bring the two fields together is as follows: In Section \ref{AML-Abstaining Machine Learning}, the paper first addresses the task of explaining the idea of AML, giving an overview of the different kinds of AML systems, and clustering the different algorithms into classes based on two dimensions. One dimension describes different reasons for abstention, i.e.,\ different situations in which an abstaining output is issued (Subsection \ref{AML-AmbiguityandOutlier}). The second dimension describes different ways in which abstention is (conceptually and technically) implemented in the system (Subsection \ref{AML-attachedandmerged}).
\\
\indent
In the second part of the paper, the philosophical analysis takes place. We will draw from insights from the philosophical literature on suspension and
demonstrate how certain types of AML systems meet the criteria for suspending judgment. First, we will draw comparisons between the reasons for abstention detailed in Part \ref{AML-Suspension-Reasons} and the various reasons (or norms) for suspension.
Secondly, in Part \ref{AML-Suspension-Nature}, we will compare the methods of implementing abstention to the nature and the forms of suspension explored in philosophy, addressing the question of which types of AML systems possibly correspond to suspension.\\
\indent
Additionally, this paper seeks to explore the broader topics within the philosophy of artificial intelligence that have not been previously applied to this specific category of machine learning systems. As our focus in this paper is on AML systems, which we have identified as a potential type of ML system capable of suspension, we will expand specific questions in the philosophy of AI to this kind of system.
In particular, we will delve into matters concerning the autonomy and explainability of machine learning-generated responses. We will apply these two questions to the abstaining output of ML systems and discuss how autonomous (Subsection \ref{AML-Autonomous}) and how explainable (Subsection \ref{AML-XAI}) the abstaining output is or can be.
We will argue that the different types of abstaining systems presented in Section \ref{AML-Abstaining Machine Learning} offer different answers for these two questions.\footnote{A more elaborated analysis, a more thorough philosophical representation on suspension and doxastic neutrality as well as an analysis of other AI systems can be found in the dissertation \citep{schuster.2024}.}
\section{Abstaining Machine Learning}\label{AML-Abstaining Machine Learning}
In this paper, we consider \emph{predicting} ML systems. In general, the task of those kinds of ML systems is to select a defined answer from an answer set $A$ for a question $Q$. The examples considered here refer to cases where the answer set $A$ is a finite, discrete set.
A familiar example is that of an image classifier. If an image classifier is to identify the breed of dog depicted in an image, the system is asked the question $Q_2=$ ``Which breed of dog is displayed in the image?'', and a possible set of defined answers is $A_2=\{\texttt{Husky}, \texttt{Labrador}, \texttt{Dachshund}, \texttt{Retriver}\}$.
\\
\indent
This type of ML is often referred to as \emph{predicting} ML and is distinct, for example, from ML in robotics, where physically acting systems are in focus, and from generative AI, where the task of the AI is to generate text, images, or other data. Moreover, the predicting systems considered here differ from other predicting systems that have a continuous, i.e.,\ infinite, set of possible answers available.\footnote{Most of the literature on AML deals with discrete classifiers. There are some studies on abstention in regression models \citep{asif.2020}, but we will not consider these here.} What is considered here is often referred to as a \emph{classifier}.\\
\indent
Moreover, we only consider so-called \emph{supervised} ML algorithms. This characteristic concerns the way the system is trained. In ML, one generally distinguishes between an application phase, in which the system solves the task that it is supposed to solve, e.g.,\ answering a question, and an earlier training or learning phase, in which the system learns how to solve the task. In the training phase, the system is equipped with some kind of training data.
Supervised systems learn to establish a relationship between the input and the desired output through \emph{labeled} training data.
For the question $Q_1$, whether a certain tumor is malignant or benign, an input data point will not consist of a whole image but of a list of measured features of the tumor, e.g.,\ its size, the number of concave points, its perimeter, and so on. The output will be the answer, i.e.,\ either \texttt{benign} or \texttt{malignant}.
In Subsection \ref{AML-Abstaining Machine Learning-Example}, we will illustrate how training data for question $Q_1$ could be visualized and provide an explanation of the mathematical properties of the training data points.
\\
\indent
When the system has learned in the training phase to connect certain questions (or lists of features) with certain correct answers (or certain labels or classes\footnote{The responses generated by a machine learning system are usually called ``outputs.'' Moreover, the terms ``label'' and ``class'' are commonly employed in literature, particularly within the context of classifiers. These terms — output, class, and label — are frequently used interchangeably. Strictly speaking, the output usually signifies the classifier's result, while the label typically refers to the ground-truth label in the training dataset. Both outputs and labels usually take up the same possible values, the values of the distinct classes. One could consider classes as abstract categories into which the data points fall. A label and an output indicate membership within one of these classes.\label{footnote-classvslabelvsoutput}}), it can later apply this knowledge in the application phase by answering new, previously unanswered questions, i.e.,\ new, unseen tumors.\\
\newline
What distinguishes abstaining classifiers from conventional classifiers is the option to choose none of the defined answers of the answer set $A$ as an output. AML can issue an \textit{abstaining output} as a response to the question $Q$ allowing an alternative to the defined answers.
Therefore, AML systems are often referred to as possibly \emph{rejecting} the task or refusing to give an answer. This rejection may be issued in the form of an output saying \texttt{I do not know}, \texttt{I abstain}, \texttt{I reject a prediction}, etc.
\\
\indent
This seems to be appropriate in many application domains. Most prominently, researchers have argued that in high-stakes scenarios like medical decision-making, ML systems with an abstaining option are clearly preferable as diagnostic tools (for example for cancer, COVID-19, or liver disease detection) \citep{kompa.2021, Brinati.2020, hamid.2017, kempt.2022}.
But also, in other application areas like weather and climate diagnostics \citep{Barnes.2021} or simple spam filters \citep{Artelt.2022a}, the abstaining option is often considered desirable.
If ML systems are to serve as expert or advice systems, it is recommended that these systems liberally admit their own uncertainty in critical situations instead of making a decision at any cost. This also corresponds to our expected behavior of human experts, as \citet[][p.~1]{Ferri.2004} point out: ``When we use human assistance for supporting decision making, there are some cases where the expert says `I don't know' and asks for further assistance (to other experts) or just prefers to postpone the decision. Frequently, we say a person is an expert or a wise person when she prefers to be silent (and ask other experts) rather than to make a mistake.'' Moreover, as \citet[][p.~292]{Campagner.2019} point out, when abstaining ML systems alert us to uncertainties, this often gives us the opportunity to improve the basis for decision-making: `` [...] because it could be used in a human in the loop setting, to point out to the human decision-maker which instances might require the acquisition of further or more precise information.''\\
\newline
\noindent
In the following, we will illustrate the domain of AML classifiers using two dimensions. Along the first dimension, we distinguish the different reasons for abstention. Thus, we give an overview of situations in which abstaining ML is in play. For this purpose, we distinguish between \textit{outlier abstention} and \textit{ambiguity abstention}.
The second dimension describes the composition of the algorithms. Here, we basically distinguish two ways in which the abstention option can be technically and conceptually integrated into an ML algorithm. We call these two types of AML systems \textit{attached} and \textit{merged abstention}. The two dimensions are fundamentally independent. One dimension concerns the reasons for abstention, and the other dimension concerns the implementation of abstention. In principle, therefore, any combination of outlier or ambiguity abstention with attached or merged abstention is possible.\\
\indent
In presenting the AML systems and their distinctions along the two mentioned dimensions, we will
revisit the question $Q_1$ concerning cancer detection and furnish an example with real-world parameters and training data points.
\subsection{An ML Example for Cancer Detection}\label{AML-Abstaining Machine Learning-Example}
\subsubsection*{The Training Data}
\begin{figure}[H]
    \centering
     \includegraphics[scale=.55]{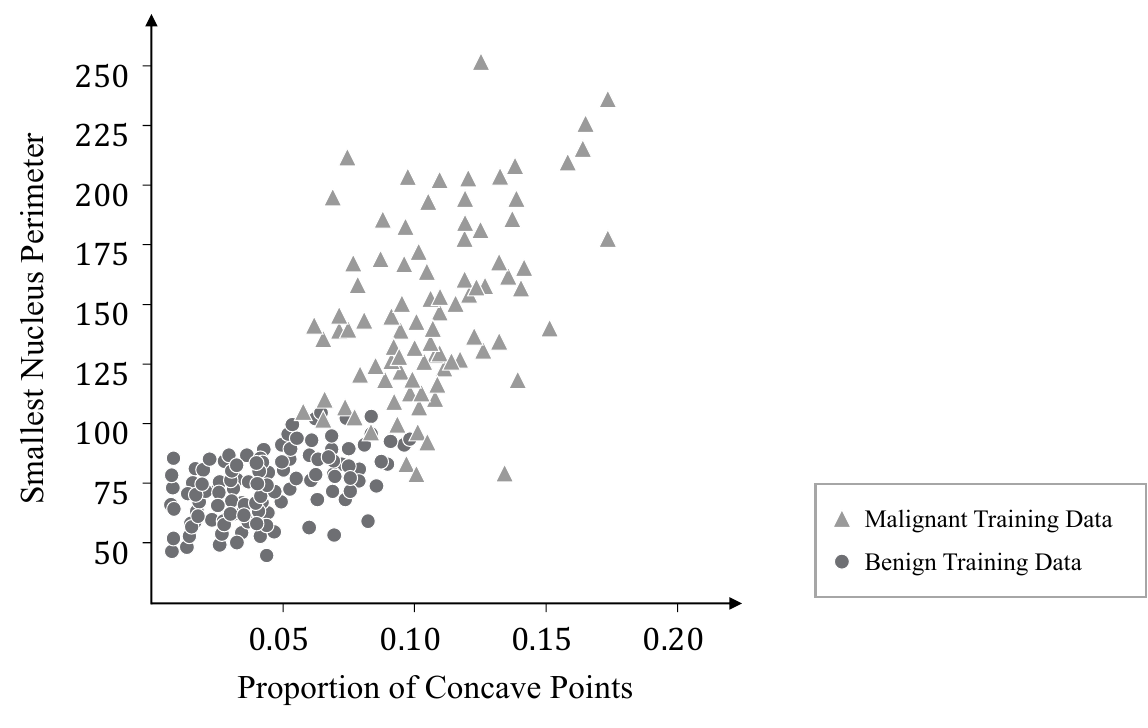}

      \caption{Training Data for Cancer Detection: Malignant data points are represented by triangles; benign data points by circles.}
         \label{AML-Figure-CancerExample}
\end{figure}

A data set for benign and malignant points that is often used can be found in \citep{breastcancer}. This data set comprises multiple features, i.e.,\ input variables, from which we have selected two (the smallest nucleus perimeter and the proportion of concave points) to visualize a two-dimensional input space.
In Figure \ref{AML-Figure-CancerExample}, an 
extract of these training data points is sketched.\footnote{The values of the visualized data points are not extracted from the data set. Rather, for this particular case study, the rough distribution of the malignant and benign data points in the data set is only sketched in order to obtain a better visualization. The range in which the data points occur is still correct.}\\

\noindent
Figure \ref{AML-Figure-CancerExample} illustrates possible training data points for training an algorithm to answer $Q_1$.
The training data points are illustrated by the circles and the triangles in the two-dimensional coordinate system in the figure. Mathematically, each training data point can be described by a tuple $\langle \bm{x}^{(i)}, y^{(i)} \rangle$, $i=1, \dots, n$.\\
\indent
In this tuple, $\bm{x}^{(i)}$ is a two-dimensional vector, which represents two input parameters: the smallest nucleus perimeter and the proportion of the concave points. For example, it could be $\bm{x}^{(i)}=(0.17, 152)$ with $0.17$ being the proportion of the concave points (ranging from $0$ to $1$) and $152$ the value for the smallest nucleus perimeter (in micrometers).
As each of the two entries of $\bm{x}^{(i)}$ is real-valued, $\bm{x}^{(i)}$ is an element of the two-dimensional real space, i.e.,\ $\bm{x}^{(i)}\in \mathbb{R}\times\mathbb{R}=\mathbb{R}^2$.
In Figure \ref{AML-Figure-CancerExample}, the value $\bm{x}^{(i)}$ is represented by the \textit{position} of the circle (or triangle) in the coordinate system, i.e.,\ by \emph{where} the circle (or triangle) lies with respect to the horizontal and vertical axis.
The space that contains all the training data points is called \emph{input space}, which is in general denoted by $X$. For our example, it is $X=\mathbb{R}^2$.\footnote{In fact, it makes sense to restrict the space of $X$ to a subset of $\mathbb{R}^2$ for this example. The proportion of concave points is measured in a value between $0$ and $1$, suggesting the interval $[0,1]\subseteq \mathbb{R}$ and the smallest nucleus perimeter is measured in micrometers suggesting to at least restrict the input space to the space of all positive-valued reals $\mathbb{R}^{+} \subseteq
\mathbb{R}$. A medical reasonable subspace would be even smaller, as the nucleus perimeter can certainly not become arbitrarily large.}\\
\newline
Since we consider supervised ML, a training data point, $\langle \bm{x}^{(i)}, y^{(i)} \rangle$, however, consists not only of the input values but also of the respective (ground-truth) label. In the breast cancer example, we not only know for a specific training data point its smallest nucleus perimeter and its proportion of concave points, but we also know whether that training data point \textit{is in fact} a malignant or a benign one. This information is stored in $y^{(i)}$. In our example case, $y^{(i)}$ can have one of the values: \texttt{malignant} or \texttt{benign}.
In Figure \ref{AML-Figure-CancerExample}, the value of $y^{(i)}$ is represented by the \emph{shape} drawn in the graph. If $y^{(i)}=\texttt{malignant}$, the point is represented by a triangle, if $y^{(i)}=\texttt{benign}$, the point is represented by a circle. The set of the potential labels is also called the \emph{output set}, as the task of the ML system becomes to predict these labels. It is in general denoted by $Y$. For our example, it is $Y=\{\texttt{malignant}, \texttt{benign}\}$, which is identical to the set $A_1$, the set of possible answers to $Q_1$.
\\
\indent
In total, one example of a training data point $\langle \bm{x}^{(i)}, y^{(i)} \rangle$ with $\bm{x}^{(i)} \in X$ and $y^{(i)}\in Y$ is always an element of the Cartesian product of the input and the output set, i.e.,\ $\langle \bm{x}^{(i)}, y^{(i)} \rangle \in X\times Y$.
For our breast cancer example, one concrete training data point could be $\langle \bm{x}^{(i)}, y^{(i)} \rangle = \langle(0.17, 152), \texttt{malignant}\rangle \in \mathbb{R}^2 \times \{\texttt{malignant}, \texttt{benign}\}$.
The complete training data set is denoted by $T$, i.e.,\ $T=\{\langle \bm{x}^{(1)}, y^{(1)} \rangle, \langle \bm{x}^{(2)}, y^{(2)} \rangle, \dots, \langle \bm{x}^{(n)}, y^{(n)} \rangle \}\subseteq X\times Y$.
\subsubsection*{The Training Phase}
As for every supervised ML classifier, the goal is to build a classifier that tells you for any arbitrary input (any vector $\bm{x}\in X$), representing a \textit{new, unseen tumor}, whether that input is benign or malignant. For this, a training phase is necessary where a connection between certain input values and the different output classes can be established, based on the given training data.\\
\indent
For example, it might be determined that a proportion of concave points above $0.15$ occurs only in malignant cases.\footnote{Commonly, these rules found by the algorithm are not that simple and are not even expressible in a way that the user or programmer would understand. Rather, they are encoded, e.g.,\ via the enormous number of parameters of a deep neural network.} This means that the algorithm tries to find a \emph{decision boundary}\footnote{If we have a multi-class problem, one boundary will not be enough.} between the different training data points that separates the data points that belong to the malignant class from the data points that belong to the benign class. An example of such a boundary can be visualized by a line in the input space, separating malignant and benign training data points.\\
\indent
Mathematically, the separation of the data points (in the input space) can be represented by a function $f$ which maps \emph{any} input vector $\bm{x}\in X$ to an output $y \in Y$. According to the above definitions, $X$ is called the input space (or set) and $Y$ is the output set of the function $f$.
How can we find such a function? We can start by considering those functions $f: X \rightarrow Y$ that use the simplest decision boundary, i.e., a line, as we will see in Figures \ref{AML-Figure-labelledAbstention} and \ref{AML-Figure-UnlabelledAbstention}. This means, we consider a linear model.\footnote{Considering only linear models is one possible \emph{model choice}. Instead, one could also make a different model choice, like a quadratic or logarithmic model, returning curved decision boundaries.
In principle, though, the set of possible functions is always restricted by a particular choice of a model, e.g.,\ to avoid overfitting or too much computational complexity, see \citet{murphy.2022}.} Overall, the possible candidate functions of a particular model choice can be collected in a set $\mathcal{F}$.
The goal is then to choose one function, to be denoted $\hat{f}$, in $\mathcal{F}$ that has the property of performing the mapping of the input parameters of \textit{the training data} in the best possible way. This means that the task in our binary classification problem is to find a $\hat{f}$ for which $\hat{f}(\bm{x}^{(i)}) = y^{(i)}$ for as many $i=1,\dots, n$ as possible. 
\\
\newline
But how can we determine $\hat{f}$ and derive a boundary that separates the training data labeled \texttt{malignant} from the training data labeled \texttt{benign} best?
One option would be \textit{to try different} functions in $\mathcal{F}$ and choose the one that makes the fewest mistakes (trial and error).
\\
\indent
The different functions in $\mathcal{F}$ then have to be evaluated in order to find the ``best one,'' i.e.,\ the one that maps the most $\bm{x}^{(i)}$ ($i=1,\dots, n)$ to their associated $y^{(i)}$.\footnote{In reality, for most applications, the optimal function has not only the objective to map as closely to $y^{(i)}$ as possible, but also to be ``simple enough'' to avoid the problem of overfitting. Therefore, the objective usually consists of one part that is to reduce the prediction error and a second part that \textit{regularizes} $f$, i.e.,\ avoids that $f$ perfectly fits the training data by being overly complex.
With this second part, one wants to ensure that the function not only maps the specific training data points well, but can also reasonably well \emph{generalize} beyond the training data. For more information about this regularization see, for example \citet{murphy.2022}.
For reasons of simplicity, we will only consider the first objective of mapping the training data as good as possible here. Moreover, as we limited the model choice to linear models, regularization is not relevant after all, as the model's complexity is restricted to linear functions.} 
We do this be determining for each $f$ in $\mathcal{F}$ how ``bad'' it is, i.e., \textit{how much loss} it produces for the different training data points.
For this, we introduce a \emph{loss function} $l$ which determines how much loss a particular function $f$ generates for each training data point.
This loss occurs when a data point is assigned a different label, according to the decision boundary set by $f$, compared to its ground-truth label from the training data.
For example, the training data point is labeled \texttt{benign}, and the label assigned by the algorithm (according to that boundary) is \texttt{malignant} (or vice versa).
\\
\indent
In general, the loss function is the heart of a learning algorithm.
It determines the loss a candidate function $f \in \mathcal{F}$ generates.
The total loss (also often referred to as ``cost'') is usually determined by summing up the single losses that occur when evaluating a training data point by the candidate function $f$.
\\
\newline 
A simple loss function could in general look like this: $l: Y\times Y\rightarrow \{0,1\},$ 
\begin{equation}
 l(y^{(i)}, f(\bm{x}^{(i)})) = \left\{\begin{array}{lll}
           1 &  \text{if } & y^{(i)} \neq f(\bm{x}^{(i)}) ,\\
            0 &  \text{if } & y^{(i)} = f(\bm{x}^{(i)}).
       \end{array}\right.
       \label{AML-Equation-normallossfunction}
\end{equation}
\noindent
Given a particular candidate function $f$, the loss function $l$ for one training data point is $0$ if the ground-truth label \textit{is equal} to the label determined by $f$ and is $1$ if the ground.truth label \emph{is unequal} to the label determined by $f$.\footnote{In accordance with Footnote \ref{footnote-classvslabelvsoutput}, this suggests a terminology in which we distinguish the ``ground-truth label'' from the ``output label,'' the latter being the label determined by $f$.} 
\\
\newline
The optimal function $\hat{f}$ is then $f \in \mathcal{F}$ for which \emph{the sum} of the values of the loss function over \emph{all} training data points $\langle \bm{x}^{(i)}, y^{(i)} \rangle$ ($i=1,\dots, n$) is \emph{as small as possible}.\footnote{The solution of such an optimization problem is often not guaranteed to be unique.}
Mathematically, we find this $\hat{f}$ by solving the following optimization problem:
$$\hat{f}= \underset{f \in \mathcal{F}}{\text{argmin}} \sum_{i=1}^{n} l(y^{(i)}, f(\bm{x}^{(i)})).$$
In the following, we will call $\hat{f}$ sometimes also the \emph{regular predictor}, to allow for a distinction from other predictors that are obtained in an abstaining setting.
\subsubsection*{The Application Phase}
Once we have found $\hat{f}$ in this way, we thereby found a model and a separation boundary, and we can \textit{apply} the ML model. The application phase can be represented in the following way: We take a new input vector $\bm{x}$ from $X$, which the system has not seen before, and put it through the ML system, i.e., the regular predictor $\hat{f}$. The output $\hat{f}(\bm{x})$ then indicates the assigned label for the input $\bm{x}$. This application phase is visualized in Figure \ref{AML-Figure-Outlier}.\\

\begin{figure}[H]
    \centering
      \includegraphics[scale=.44]{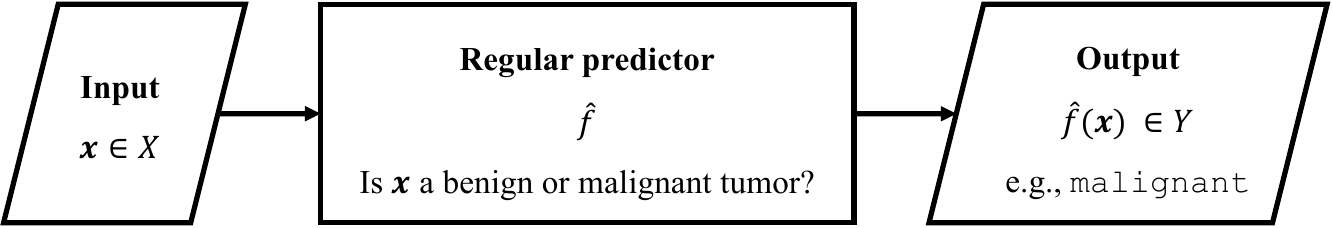}
      \caption{Flowchart of the application phase of a regular (non-abstaining) ML classifier: The input $\bm{x}$ is processed through the regular (non-abstaining) predicting function $\hat{f}$ and an output $\hat{f}(x)$ from the output set $Y$ is generated.}
         \label{AML-Figure-Outlier}
\end{figure}
\noindent
The real-world example for question $Q_1$ will be revisited and applied in the next two subsections when introducing the domain of AML classifiers, highlighting the two differentiations between ambiguity vs.\ outlier abstention (Subsection \ref{AML-AmbiguityandOutlier}) and attached vs. merged abstention (Subsection \ref{AML-attachedandmerged}).
\subsection{Reasons for Abstention: Ambiguity versus Outlier Abstention}\label{AML-AmbiguityandOutlier}
The first distinction in abstaining machine learning revolves around the \textit{reasons} prompting a system to abstain. This distinction describes the handling of a \textit{new} data point during the \textit{application phase} of an AML algorithm. Therefore, the following elaborations have to be considered at a stage where the system is already trained and is applied to new data points.\\
\indent
In general, if it is too uncertain whether the system will produce the correct output for the new data point, an AML system will abstain. This uncertainty can arise in many ways. While some uncertainties concern the general structure of the model (e.g.,\ an inappropriate model choice for the kind of training data), other uncertainties are due to some characteristic of a specific input.\\
\indent
The different uncertainties can be categorized by means of a common distinction in abstaining machine learning: the distinction between ambiguity and outlier abstention.\footnote{Uncertainties are commonly categorized into aleatoric and epistemic \citep{derKiureghian.2009, hullermeier.2021}. Aleatoric uncertainty arises from inherent randomness or statistical variability, while epistemic uncertainty stems from a lack of knowledge. Consequently, epistemic uncertainty is generally considered reducible, whereas aleatoric uncertainty is not. Although this paper primarily focuses on the downstream characterization of outlier and ambiguity abstention, the distinction between aleatoric and epistemic uncertainty remains relevant. \citet{hullermeier.2021} argue that outlier abstention typically reflects epistemic uncertainty, as it is associated with missing information (e.g., insufficient training data) in the outlier region. On the other hand, abstention models based on ambiguity are more closely linked to aleatoric uncertainty.} Roughly speaking, when an input is too far away from or too dissimilar to the training data, we are dealing with an outlier; when the input is such that more than one output is likely for the input, we are dealing with ambiguity.
This distinction can be found in early works \citep{dubuisson.1993, Denoeux.1995} and is sometimes referred to with different names, such as novelty rejection versus ambiguity rejection \citep{hendrickx.2021}, distance rejection versus ambiguity rejection \citep{dubuisson.1993} or distance rejection versus confusion rejection \citep{Mouchere.2006b}.
\subsubsection*{Outlier Abstention}
In outlier abstention \citep{Lotte.2008, Mouchere.2006a, Mouchere.2006b}, the system abstains on data points that are very dissimilar to the training data. This is useful for (at least) two scenarios. First, if an input is very far away from \textit{all} training data points, it is likely that the input might belong to none of the classes that are in the scope of the classifier. If a classifier is trained to classify different breeds of dogs and the new input is an image of a cat, the cat image will likely be very dissimilar to \textit{all} of the different dog images that were used for training the classifier. The classifier here really should abstain, as it is only capable of classifying dogs and will not be able to solve the task of classifying a cat.
The correct answer for this input of a cat image (and for the question about what is displayed in the image) is not included in the set of defined answers $A_2=\{\texttt{Husky}, \texttt{Labrador}, \texttt{Dachshund}, \texttt{Retriver}\}$ that the system operates on. Hence, it is reasonable that the algorithm chooses none and abstains.
\\
\indent
Secondly, even in cases where the correct label of an input might be one of the considered labels of the classifier, i.e.,\ the correct answer to the question is one of the defined ones, outliers appear. 
If an input dog image is very dissimilar to the training images, this suggests that any prediction the system could make will be prone to error.
The data point can be dissimilar to the training data for various reasons: There could be measurement inaccuracies, there could be adversarial examples (that are meant to trick the system), or the training data have been just not diverse enough \citep{hendrickx.2021}.
In this sense, outlier detection is often used to actually improve the prediction system. If a certain dog image is characterized as an outlier (although the system should recognize the type of dog in the image), this might suggest that the system was trained on too uniform and not sufficiently diverse data, which could be improved based on the detected outliers. Maybe the system was trained on images of dogs that were taken during summertime and the detected outlier is a dog image in the snow. Detecting this outlier can suggest retraining the system with more diverse data; in this case: images taken in different seasons.
\newline
Figure \ref{AML-Figure-Outlier} illustrates a typical case of outlier abstention.
Similar to Figure \ref{AML-Figure-CancerExample}, the triangles represent the training data with the label \texttt{malignant} and the circles represent the training data with the label \texttt{benign}. Besides the training data, an additional data point is represented by a star.
The star represents a to-be-classified new data point that is taken to be an outlier.
\begin{figure}[H]
    \centering
      \includegraphics[scale=.55]{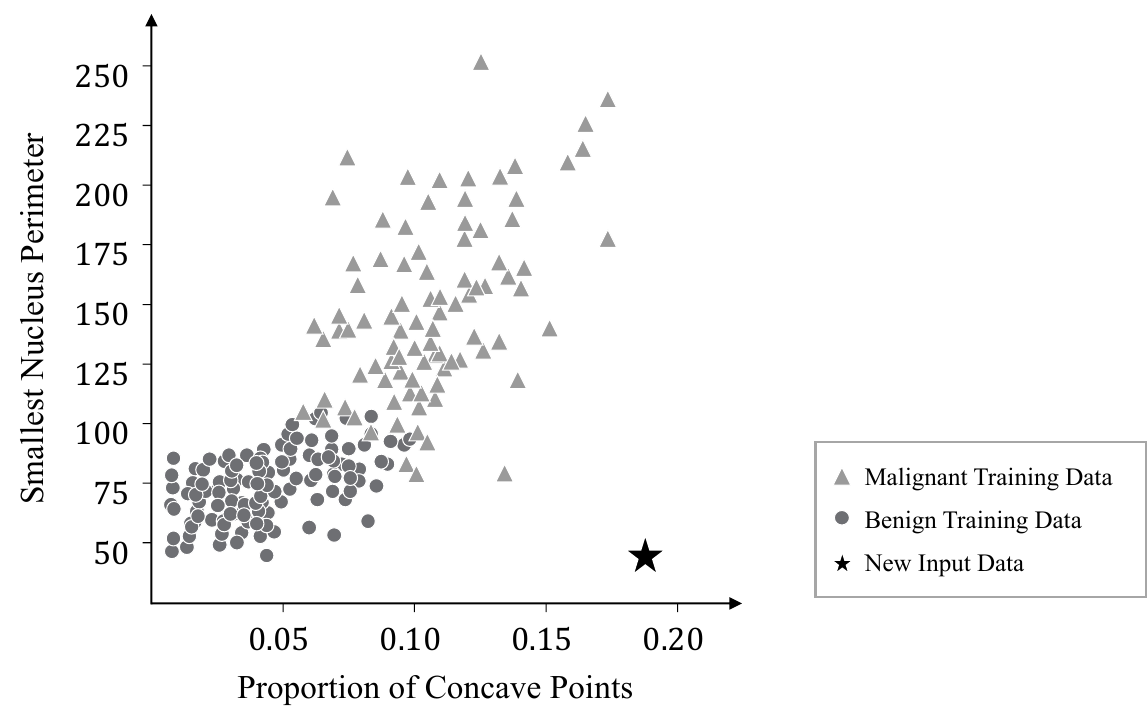}
      \caption{Outlier Abstention: A to-be-classified data point (star) is too dissimilar to training data (circles and triangles).}
         \label{AML-Figure-Outlier}
\end{figure}
\noindent
\subsubsection*{Ambiguity Abstention}
In contrast to outlier abstention, the problem in ambiguity abstention is not that none of the answers seem likely, but rather that too many of the answers seem likely for the input \citep{Barnes.2021, Campagner.2019, Sarker.2020, thulasidasan.2019}. Ambiguity is at play when an input appears to belong to more than one class. This can be the case when the input is on a boundary, but also can be due to the structure of the training data itself.\footnote{In the latter case, the uncertainty is not purely due to some characteristic of the input sample but also due to the composition of the training data being not perfectly separable or the model choice being inappropriate to perfectly separate the data.} Often training data is not perfectly separable. When this is the case, the training data is called \emph{noisy}. This means that there are certain regions in the training data that overlap (see Figure \ref{AML-Figure-Ambiguity}). 
If an input sample lies in such an overlapping (or noisy) region, ambiguity is present and a prediction for one class or the other would be error-prone. This type of uncertainty can also arise for a variety of reasons.
Maybe the input data point simply has certain characteristics of one class as well as characteristics of another class. For example, the size of the dog in an input image might be indicative of a retriever, while the coat color is clearly indicative of a Labrador.\\
\newline
A case for ambiguity abstention for the Example from Subsection \ref{AML-Abstaining Machine Learning-Example} can be visualized like this:
\begin{figure}[H]
    \centering
      \includegraphics[scale=.55]{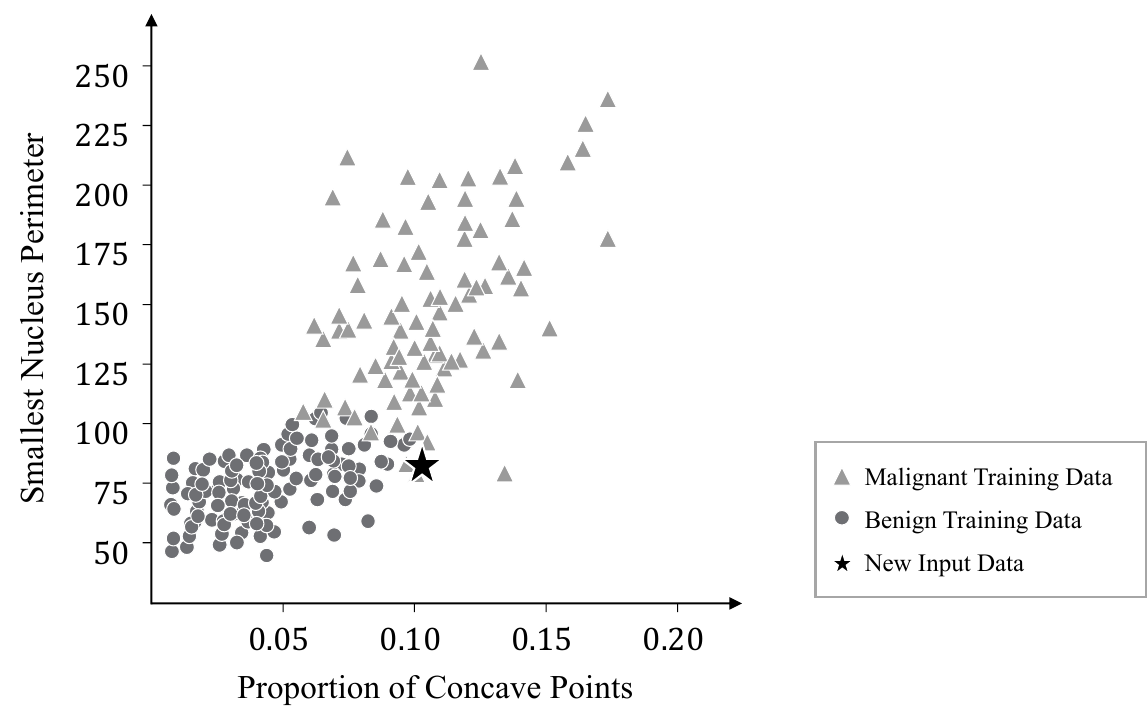}
      \caption{Ambiguity Abstention: A to-be-classified data point (star) lies in an overlapping, ambiguous area of the training data.}
         \label{AML-Figure-Ambiguity}
\end{figure}
\noindent
One important distinction between outlier and ambiguity abstention lies in how a data point can be identified as an outlier or an ambiguous point. Detecting an outlier typically does not require any information about the \textit{labels} of the training data points. As illustrated in Figure \ref{AML-Figure-Outlier}, the outlier could be identified without distinguishing between the triangle-shaped and circle-shaped training data points. The only essential information is the input values of the training data points (i.e., \emph{where} they are located in the two-dimensional space) and the input value of the new data point. The labels $y^{(i)}$ of the training data are not needed.
\\
\indent
In contrast, to identify a new data point as an ambiguous case, information about the labels of the training data is essential (i.e., the information $y^{(i)}$ is necessary). Furthermore, determining whether a new data point $\bm{x} \in X$ is an ambiguous case often depends on the specific trained model and cannot be directly inferred from the training data and $\bm{x}$ alone. While the potentially ambiguous region is visually discernible in Figure \ref{AML-Figure-Ambiguity}, this is not always the case, especially not for higher-dimensional data and more complex models. This consideration is picked up again in the distinction between two forms of attached abstention, as discussed in Section \ref{AML-attachedandmerged}.
\subsection{Implementation of Abstention: Attached versus Merged Abstention}\label{AML-attachedandmerged}
In this section, we introduce the second dimension for classifying AML systems. Here, we distinguish different \textit{types} of AML systems with respect to the technical implementation of the abstention option. Although there are many ways to incorporate the abstention option into a classifier, we will present two main categories under which many systems can be subsumed and that we consider to be fundamentally different approaches.
In contrast to many other reasonable approaches to categorizing different abstaining models (see especially \citet{hendrickx.2021}\footnote{Note that a new version of \citep{hendrickx.2021} is published as \citep{hendrickx.2024}. This paper refers to the previous version though.}), our distinction between attached and merged abstention models is chosen for being most relevant and useful for the philosophical questions considered in Section \ref{AML-Philosophy}. In Section \ref{AML-Philosophy}, we will see that the different types of abstaining models behave differently regarding the questions about their similarity to suspension, their autonomy, and their explainability.
\subsubsection*{Attached Abstention}
\begin{figure}[H]
    \centering
      \includegraphics[scale=.44]{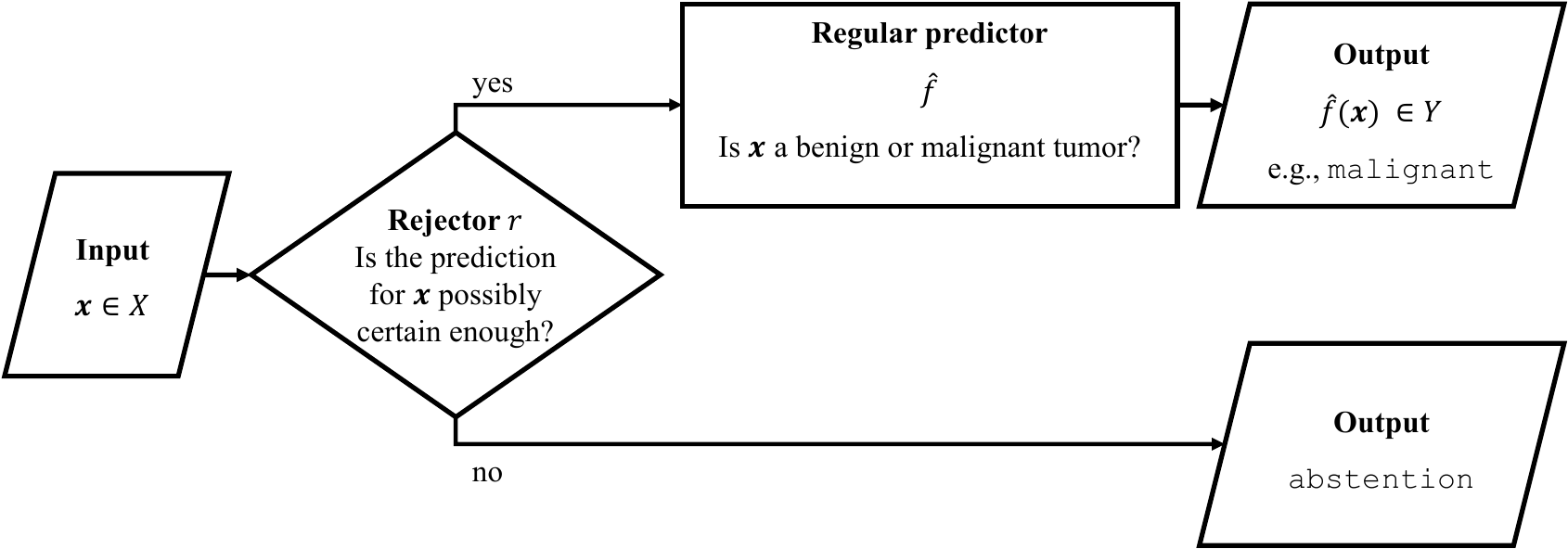}
      \caption{Pre-algorithmic attachment of abstention.}
         \label{AML-figure-preattachment}
\end{figure}

\begin{figure}[H]
    \centering
      \includegraphics[scale=.44]{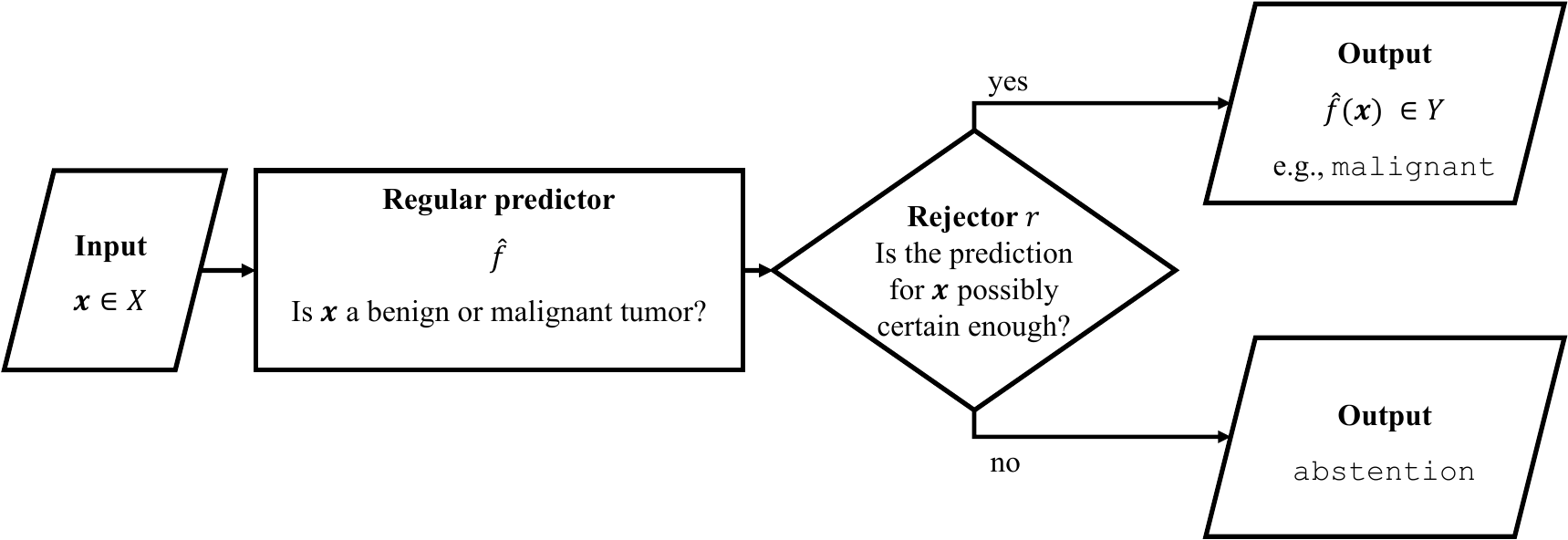}
      \caption{Post-algorithmic attachment of abstention.}
         \label{AML-Figure-postattachment}
\end{figure}

The first class we will consider is the class of what we will call attached abstaining machine learning systems. In these systems, the part that is relevant for the abstaining activity is in some sense \textit{attached} to the core machine learning algorithm, i.e.,\ to the predicting algorithm \citep{Sarker.2020, Mouchere.2006b}. Hence, the predicting and the abstaining activities are separated from each other and one can speak about ``the predictor'' (which we refer to as $\hat{f}$) and ``the rejector,'' $r$ (i.e.,\ the part of the system that is relevant for abstaining). There are two ways in which the rejector can be attached to the predictor. The rejector can be attached \emph{prior} or \emph{posterior} to the predictor.
\begin{itemize}
\item[(a)] \emph{Pre-algorithmic attachment}\\
In pre-algorithm abstention models, the abstaining part is executed prior to the predicting classifier\footnote{What \citet{hendrickx.2021} call a ``separated rejector'' can best be compared to pre-algorithm abstention models.} \citep{wu.2007, Mouchere.2006a, homenda.2014, coenen.2020}.
This means that that for a given input, the rejector decides whether or not to abstain for the input even \textit{before} the prediction algorithm starts. If the input is not rejected, the predictor starts running; if the input is rejected, the predictor will not even be started in the first place.
\\
\indent
Pre-algorithmic abstention is especially relevant for outlier abstention \citep{coenen.2020, Lotte.2008}. For a given input, the decision of whether the prediction will be too uncertain is made before the prediction is computed. Therefore, it must be a property that is inherent to the input data that determines whether the input will be rejected. This does not work well for ambiguity rejection because ambiguity arises not only due to the input but due to the relationship of the input and the trained model.
The concept of pre-algorithmic attachment is visualized in Figure \ref{AML-figure-preattachment}.
\end{itemize}
\noindent
\begin{itemize}
\item[(b)] \emph{Post-algorithmic attachment}\\
For post-algorithmic abstention, the rejector is downstream of the predictor \citep{Campagner.2019, Brinati.2020, Artelt.2022a}. For every input data point, an ordinary prediction is calculated. This is done independently of any abstention activity. The prediction is computed in the exact same way the prediction would be computed in a non-abstaining system. This means that the question that is under discussion, $Q$, is answered by choosing one of the defined answers from $A$. In the second step, the certainty of the prediction, i.e.,\ the likelihood of the selected defined answer being the correct answer is measured. This certainty can be provided by the predictor itself (e.g.,\ as some kind of probability value in a neural network, distance in a support vector machine, or some ``soft probabilistic classifier'' \citep{Campagner.2019, Brinati.2020})
or it can be calculated additionally by some uncertainty or reliability measure \citep{linusson.2018, Mouchere.2006a, Lotte.2008}.
This certainty value is then used in the posterior attached rejector. In the simplest version, the rejector only consists of a certainty threshold and two \emph{if}-clauses. If the certainty of the calculated answer being correct is above the threshold, the prediction is passed through and revealed; if the certainty is below the threshold, the predicted answer is rejected, and the system abstains.\footnote{Although the abstaining part of this type of model is attached, it corresponds best to what is called a ``dependent rejector'' in \citet{hendrickx.2021}.
The term ``dependent rejection'' used by \citet{hendrickx.2021} implies that the rejection of a particular input \textit{depends} on the previously calculated output of the predictor. This stands in contrast to what we refer to as the (attached) \textit{pre-algorithmic} abstention models, wherein the rejection of an input occurs prior to the predictor's calculation and is, in that sense, \textit{independent} of the predictor.} The concept of post-algorithmic attachment is visualized in Figure \ref{AML-Figure-postattachment}.
\end{itemize}
\noindent
Both pre-algorithmic and post-algorithmic attachment are attached forms of abstention since the abstaining part is in both forms an additional, separated algorithm that is attached (either prior or posterior) to the predictor.
Attached abstention could also be called \emph{threshold abstention} as the decision whether a sample is rejected or not is usually based on comparing some certainty (in the case of post-algorithmic abstention) or similarity (in the case of pre-algorithmic outlier abstention) to a defined threshold, see \citet{hendrickx.2021}.\footnote{Note that there are varieties of attached AML systems that do not include a pre-set certainty threshold. For example, it is possible to reject a fixed fraction of the samples.
In this approach, it is not a matter of rejecting all samples below a specific \textit{certainty threshold}; instead, a \textit{fixed fraction} of the most uncertain samples, for instance, the bottom $10\%$, is rejected.}
\subsubsection*{Merged Abstention}
The crucial difference between merged and attached AML systems is that for the merged systems the abstaining and predicting activity are to some extent inseparable. The abstaining activity is neither upstream nor downstream of the prediction but is included in the predicting activity.
Therefore, it is not practical anymore to refer to ``the predictor'' and ``the rejector.'' Instead, the predictor is modified to have the capability to reject as well. For merged AML systems, we can aptly name the modified predictor an ``abstention predictor.''\\
\indent
In a classifier, an extra, abstaining output is introduced. In addition to the outputs represented by the defined answers, there is also the abstaining output. For a given input (e.g.,\ a dog image), the system can either output one of the defined answers (e.g.,\ \texttt{Husky}, \texttt{Labrador}, etc.)\ or output the \texttt{abstention} output.
\\
\indent
The property of being ``merged'' can be observed both in the application phase and in the learning or training phase of the algorithm.
In the application phase, the fact that the AML system is ``merged'' is illustrated by the fact that decisions about whether to abstain on an input are made neither before nor after the decision about which output to assign (if any). The decision about abstention is made simultaneously with, and as part of the decision about the appropriate output.
In the application phase, \texttt{abstention} is simply one additional output among others and in this sense one additional answer.
For this, we do not use the regular predictor $\hat{f}$, but a special abstention predictor $\Bar{f}$, which also allows for abstention.
The application phase of a merged AML system can be visualized in the following flowchart:\\
\begin{figure}[H]
    \centering
      \includegraphics[scale=.44]{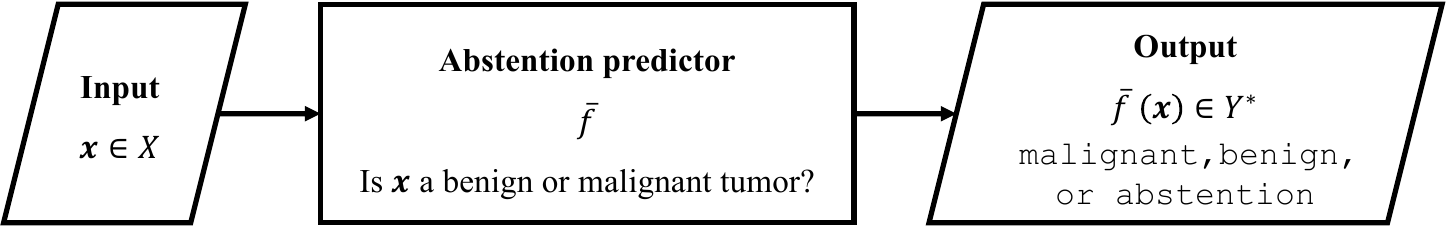}
      \caption{Merged Abstention: The decision about abstaining or not is made simultaneously to the decision about the class by an adapted abstention predictor $\Bar{f}$.}
         \label{AML-Figure-Merged}
\end{figure}
\noindent
In order to obtain such an abstention predictor $\Bar{f}$, the training phase of a merged AML system has to be adapted.
Those adaptions in the training phase, i.e.,\ the way in which the abstaining option is learned, illustrate the second dimension in which merged AML systems differ from attached AML systems. For merged AML, the tasks of rejecting and predicting are blended into one task that is \textit{learned simultaneously} in the training phase.\footnote{This is also why \citet{hendrickx.2021} call this type of learning \emph{simultaneous learning} as contrasted with sequential learning.}
While it is possible for an attached AML to have the same learning phase as a non-abstaining classifier, the learning phase of a merged AML is necessarily different from a non-abstaining classifier.
\\
\newline
With Labeled Abstention (a) and Unlabeled Abstention (b), we will distinguish again between two ways of how the learning phase of a merged AML system can allow for abstention-learning. This distinction concerns only the training phase and the way the abstaining class is learned.
\\
\indent
We will explain this by means of the cancer detection example from Subsection \ref{AML-Abstaining Machine Learning-Example}. There, we introduced how a \emph{regular, non-abstaining} classifier $\hat{f}$ can be trained on the training data visualized in Figure \ref{AML-Figure-CancerExample}.
This training or learning phase can now, in principle, be adapted in two ways in order to allow for abstention.
\begin{itemize}
    \item[(a)] \textit{Labeled Abstention}\\
A simple solution for training a system when to abstain is to extend the general method of supervised learning from the normal outputs to the abstaining output. In the training phase, a classifier is usually given examples of inputs (e.g.,\ images of dogs) along with the correct (ground-truth) label or output we want for that particular image. For the dog classifier, in the training phase, the system would be presented with multiple images of huskies all labeled \texttt{Husky}, multiple images of retrievers all labeled \texttt{Retriever}, etc.
The system is shown what a conventional input of a dog image looks like, for which we want to have \texttt{Retriever} as the output.
Analogously, we can now proceed for the abstention class. One can label inputs for which one would consider abstention appropriate with the label \texttt{abstention} and put them into the training phase just like the examples of all other classes \citep{Lotte.2008, Mouchere.2006a, singh.2004}.\footnote{This need not to be the end result of training the classifier. In \citet{singh.2004}, the authors use the rejected training data to retrain the classifier with potentially new classes earlier detected as outliers.
}
For example, one could label images of Shepherds, Bulldogs, or images of cats by hand with \texttt{abstention} since these images should be considered outliers. Moreover, blurry images or images where the dog is only partially visible can also be labeled \texttt{abstention} by hand.
Thereby the set of defined answers is in a sense extended from $\{\texttt{Husky}, \texttt{Labrador}, \texttt{Dachshund}, \texttt{Retriver}\}$ to $\{\texttt{Husky}, \texttt{Labrador}, \texttt{Dachshund}, \texttt{Retriever}, \texttt{abstention}\}$.\\
\newline
Considering the example in Figure \ref{AML-Figure-CancerExample}, in the original, non-abstaining case, a training data point was a tuple $\langle \bm{x}^{(i)}, y^{(i)}\rangle$ with $\bm{x}^{(i)}\in X=\mathbb{R}^2$ and $y^{(i)}\in Y=\{\texttt{malignant}, \texttt{benign}\}$.
In the case of labeled abstention, some of the training data points have the label \texttt{abstention}, i.e.,\ $y^{(i)}=~$\texttt{abstention}. Hence, for a training data point $\langle \bm{x}^{(i)}, y^{(i)}\rangle$, it is $y^{(i)}\in Y^{*}$ with $Y^{*}=\{\texttt{malignant}, \texttt{benign}, \texttt{abstention}\}$.
The training data for this would be the set $T^{*}=\{\langle \bm{x}^{(1)}, y^{(1)} \rangle, \langle \bm{x}^{(2)}, y^{(2)} \rangle, \dots, \langle \bm{x}^{(n)}, y^{(n)} \rangle \}\subseteq X\times Y^{*}$.
\\
\indent
In this approach, there is no categorical change required for the loss function. The loss function only needs to be extended to accommodate the extra class. The loss function for the non-abstaining, binary classification from Equation (\ref{AML-Equation-normallossfunction}) is a function from $Y\times Y$ to the loss $\{0,1\}$. A loss function for the labeled abstaining case can be the same as $l$, only mapping from the extended sets, i.e., from $Y^{*}\times Y^{*}$. The training data for labeled abstention is visualized in Figure \ref{AML-Figure-labelledAbstention}.\\

\begin{figure}[H]
    \centering
      \includegraphics[scale=.55]{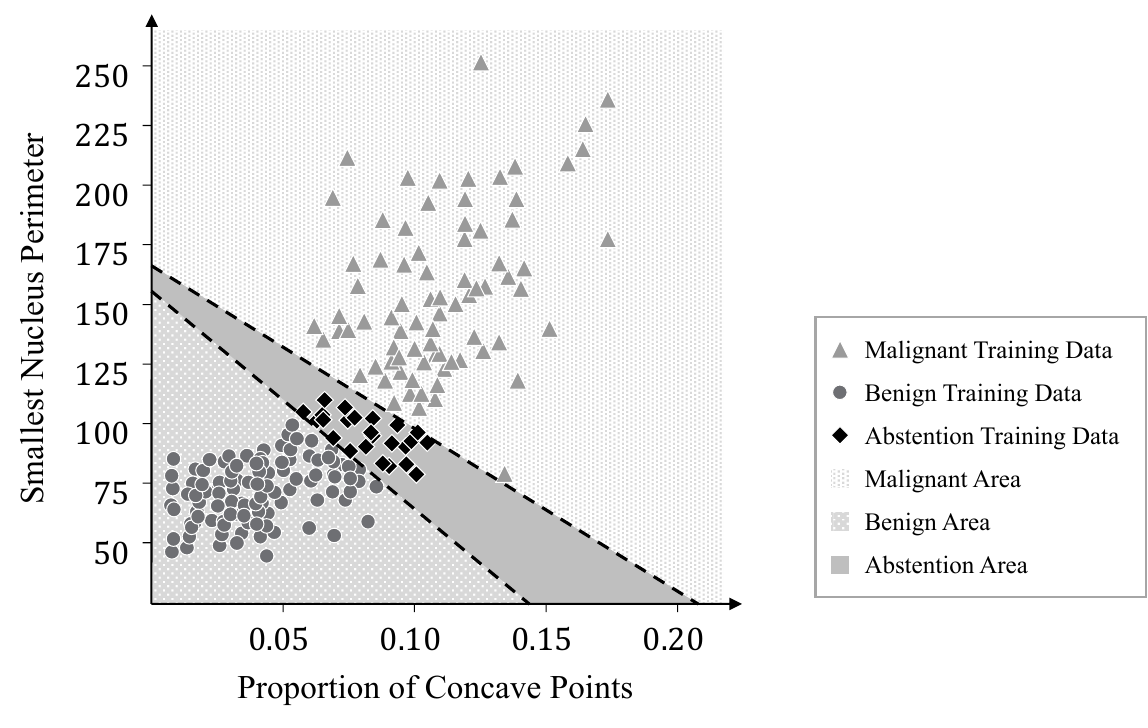}
      \caption{Labeled Abstention: Training data is either labeled \texttt{malignant} (triangle), \texttt{benign} (circle), or \texttt{abstention} (diamond). The model learns three different areas for the three different classes: a malignant area (top), a benign area (bottom), and an abstaining area (middle).}
         \label{AML-Figure-labelledAbstention}
\end{figure}
\noindent
In this classification problem, simply three classes instead of two are considered.
This means that the model needs to learn two boundary lines instead of just one as evident in Figure \ref{AML-Figure-labelledAbstention}.
\\
\newline
This approach has two major drawbacks, though.
First, labeling training data points with \texttt{abstention} by hand can be very time-consuming.
Second, often it is not useful to label training data as \texttt{abstention}.
While in some application domains, we know exactly what a prototypical abstention case might look like (e.g.,\ a blurred image for an image classifier), often we do not, or at least not in advance. In particular, when the uncertainties are due to factors that cannot be readily detected by humans looking at the training data, we cannot tell which samples will be error-prone. Often, the samples that are difficult for the algorithm to process are easy for a human expert and vice versa. This suggests that the human expert will not be able to identify the difficulties for the machine, so that it is unclear how the abstention labels are determined in the training data.\\
\item[(b)] \textit{Unlabeled Abstention}\\
Besides the straightforward way of inserting abstention as an extra output in the learning process as described in case (a), there is a more indirect, but also more sophisticated way. Here, the training data is not explicitly labeled \texttt{abstention}. In systems like those of \citet{thulasidasan.2019, Geifman.2019, mozannar.2020, wegkamp.2011, Barnes.2021, Yuan.2020}, the training data looks exactly the same as in a training situation of a \textit{non-abstaining} classifier. There are images of the different dog breeds, and each image is labeled with one of the normal (defined) labels, i.e.,\ \texttt{Husky}, \texttt{Retriever}, \texttt{Dachshund}, or \texttt{Labrador}. No training image has the label \texttt{abstention}.
Hence, for our main working example from Subsection \ref{AML-Abstaining Machine Learning-Example}, the set of training data for the unlabeled abstention case would be $T=\{\langle \bm{x}^{(1)}, y^{(1)} \rangle, \langle \bm{x}^{(2)}, y^{(2)} \rangle, \dots, \langle \bm{x}^{(n)}, y^{(n)} \rangle \} \subseteq X\times Y$ with $\bm{x}^{(i)}\in X=\mathbb{R}^2$ and $y^{(i)}\in Y=\{\texttt{malignant}, \texttt{benign}\}$.
\\
\newline
Therefore, the usual supervised way in which an ML system learns to associate an input with a desired output is not applicable to the abstention cases.
In order for the system to learn a connection between certain images and the abstention output, the underlying learning process, i.e.,\ the loss function itself must be adjusted.\footnote{In \citet{hendrickx.2021}, the authors present another approach to learning to abstain and predict in what they call a ``simultaneous learning'' way. This does not require labeling the input data or directly adjusting the loss function. This workaround is usually based on combining different algorithms, each of which executes only one predicting task. For example, if there are four ordinary classes, i.e.,\ four defined answers, one could train four different classifiers in a ``one vs.\ all'' training. This can, for example, be implemented via several support vector machines (SVM), as it is done in \citet{wu.2007}. The combination of the four trained SVMs then possibly yields areas of overlap or areas that none of the classifiers considers to belong to its trained class. These areas can then be seen as abstaining areas. In our framework, we do not consider these types of algorithms to be merged systems, though. Although they do not perfectly fit the prototype of attached systems either, abstaining and predicting still happen in different parts of the algorithm. Plus, the systems do not really learn what abstaining cases look like. This will become relevant for our considerations in Subsection \ref{AML-XAI}.}
\\
\indent
This can be implemented when for a given training data point, it is possible not only to produce a full loss (if the point is misclassified) or no loss (if the point is classified correctly), but also a small loss if the point is not classified at all. 
For the breast cancer classifier, the normal (non-abstaining) loss function of Equation (\ref{AML-Equation-normallossfunction}) was introduced as a function that takes the value $1$ for each misclassified data point and the value $0$ for each correctly classified point. The abstaining loss function could then include an additional loss of, say, $0.2$ if the system does not classify \texttt{benign} or \texttt{malignant} but instead chooses the \texttt{abstention} output for a given input (regardless of what the point's actual ground-truth label is).\footnote{\label{Footnote-differentalphas}Depending on the context, it might actually make sense to assign different penalties for abstaining for different ground-truth labels. In our example, it might make sense to rate ``false negatives'' worse than ``false positives.'' Consequently, abstention for benign cases could be penalized more than abstention for malignant cases \citep{zheng.2011}.}
\\
\newline
In the case of unlabeled abstention, we look for $\Bar{f}$ in the set of the candidate functions $\mathcal{F}^{*}$, which consists of functions of a particular model choice that maps from $X=\mathbb{R}^2$ to $Y^{*}=\{\texttt{malignant}, \texttt{benign}, \texttt{abstention}\}$.
While the set of the candidate functions was also $\mathcal{F}^{*}$ for the case of labeled abstention, in unlabeled abstention training, the loss function $l^*$ needs to be adjusted, too. For each single training data point $\langle \bm{x}^{(i)}, y^{(i)}\rangle$, $l^{*}(y^{(i)}, f(\bm{x}^{(i)}))$ can add either a loss of $1$ for misclassification, a loss of $0$ for correct classification, or a loss of some $\alpha$ if the system abstains on this point. Hence, $l^{*}: Y\times Y^{*} \rightarrow \{0,1,\alpha\}$,
\begin{equation}
       l^{*}(y^{(i)}, f(\bm{x}^{(i)})) = \left\{\begin{array}{lll}
           1 &  \text{if } & y^{(i)} \neq f(\bm{x}^{(i)}) \text{ and } f(\bm{x}^{(i)}) \neq \texttt{abstention}, \\
           \alpha &  \text{if } & f(\bm{x}^{(i)}) = \texttt{abstention}, \\
            0 &  \text{if } & y^{(i)} = f(\bm{x}^{(i)}).
        \end{array}\right.
               \label{AML-Equation-abstaininglossfunction}
\end{equation}
\noindent
Note that $\alpha \in (0,1)$ since for $\alpha \leq 0$ the system would always abstain and for $\alpha \geq 1$ never abstain. If the same $\alpha$ is chosen for all classes, it has been noted in \citet{ramaswamy.2018} that $\alpha \leq \frac{m-1}{m}$ for $m$ being the cardinality of $Y$, the number of possible ground-truth labels.\footnote{This can be seen following Chow's rule for an optimal abstention rate \citep{chow.1970}. According to this rule, Equation (\ref{AML-Equation-abstaininglossfunction}) states that the system should abstain iff the probability of the likeliest output is smaller than $1-\alpha$. Note that this is only one \textit{necessary} upper bound for $\alpha$. If the prior probabilities for the different classes are highly unequally distributed, $\alpha$ should be bounded even more. In fact, in this case, considering different $\alpha$ values for the different classes is reasonable as noted in Footnote \ref{Footnote-differentalphas}.} In our example, $m=2$. This means that choosing to abstain has to be always less costly than making a random guess for a particular point. The closer $\alpha$ is to $0$, the less it costs for the system to abstain, i.e.,\ the more the system will abstain. If $\alpha$ is close to $\frac{m-1}{m}$, the system will learn to abstain only rarely, since abstention is almost as costly as making a random guess.\\
The distinction between $l$ and $l^{*}$ shows the principle of how a loss function can be adapted to allow the system to learn abstaining. It should be noted that this is a simplified loss function used for illustrative purposes. The loss functions in the literature are more complicated and designed to be handled numerically well \citep{thulasidasan.2019, Thulasidasan.2019b, Geifman.2019, Yuan.2020, Barnes.2021}.
\\
\newline
In Equation (\ref{AML-Equation-abstaininglossfunction}), we see that the option to abstain is \textit{merged} into the loss function $l^{*}$ and thereby merged into the training of the classifier. Predicting and abstaining are trained at the same time. A trained unlabeled classifier is illustrated in Figure \ref{AML-Figure-UnlabelledAbstention}. In contrast to this, attached AML systems can only learn in a sequential way. First, for example, it is learned how to classify and only then it is learned how to abstain. Moreover,  the prototypical systems of attached AML systems that we presented here do not even \emph{learn} to abstain but are rather \emph{told} by the programmer when they should abstain. 
\end{itemize}
\noindent
\begin{figure}[H]
    \centering
      \includegraphics[scale=.55]{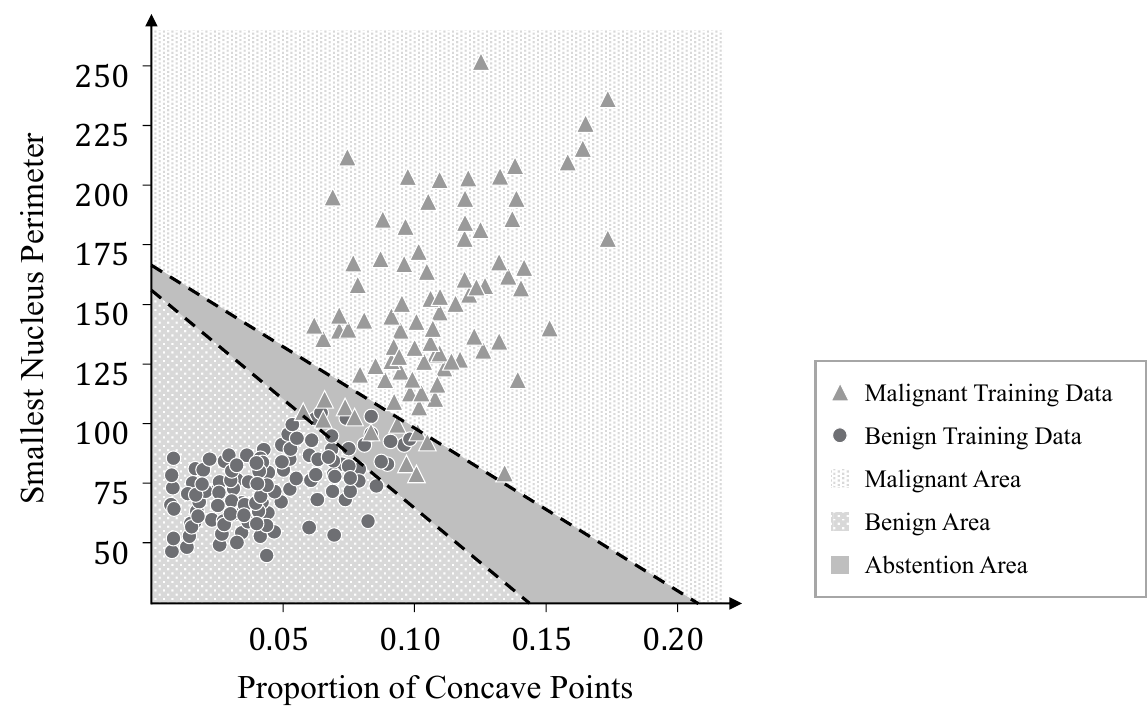}
      \caption{Unlabeled Abstention: the training data consists only of malignant (triangle) and benign (circle) points. Due to the adaption of the loss function, the model learns to separate three areas:
      a malignant area (top), a benign area (bottom), and an abstaining area (middle).}
         \label{AML-Figure-UnlabelledAbstention}
\end{figure}
\noindent
\section{Philosophical Analysis}\label{AML-Philosophy}
\subsection{Comparison of Suspension and Abstention}\label{AML-Suspension}
In the following section, we want to investigate how far the phenomenon of abstention, described in the previous section, matches its epistemological counterpart: suspension of judgment. Here, we draw parallels
between abstention and suspension, but also stress the points where the analogy ends. First, we compare the reasons to abstain that were presented in Subsection \ref{AML-AmbiguityandOutlier} with reasons to suspend (Part \ref{AML-Suspension-Reasons}), and second, we compare the different ways abstention is implemented, which was investigated in Subsection \ref{AML-attachedandmerged}, with different forms of suspension (Part \ref{AML-Suspension-Nature}).\\
\indent
In philosophy, the doxastic concepts of belief and disbelief are characterized by taking one of the defined (or complete) answers to a question to be true. Suspension is characterized by not choosing or not committing to the truth of any of the defined or complete answers, i.e.,\ being neutral towards the defined answers.
\subsubsection{Reasons for Suspension and Abstention}\label{AML-Suspension-Reasons}
The first question that is asked from an epistemological, normative point of view is: How can suspension be justified, i.e., what are the situations in which it is rational to suspend judgment towards a proposition (or question)?\footnote{Philosophers have argued that suspension tends to be question-directed rather than directed at a proposition, see, in particular, \citet{friedman.2013question}. Still, when compared to (dis)belief, it is sometimes useful to use the phrase ``suspension about a proposition $p$.''}\\
\indent
Interestingly, suspension offers a more complex normative profile than belief and disbelief do. While belief and disbelief can only be justified \emph{positively}, suspension can, according to \citet{zinke.2021}, be justified in two ways: \emph{positively} and \emph{privatively}.\\
\newline
\noindent
A justification or a reason for a belief in $p$ is always some sort of positive evidence for $p$. I am positively justified in believing that the dog is a Husky, because the dog is white or because the dog has blue eyes. All these reasons provide us with positive evidence for believing $p$. 
In some cases, we are positively justified in suspension, too. Prototypical cases are cases of vagueness \citep{ferrari.2022} or chance \citep{feldman.2018, zinke.2021}. We might suspend about the proposition ``This cup is blue'', because the cup is a borderline case between being blue and green, or we might suspend about ``This is the winning lottery ticket’’, because the lottery is fair and it is up to chance.\\
\indent
In the most prototypical cases, though, suspension is justified differently. Usually, we do not suspend because we have positive evidence \emph{for} suspension, but because we do not have enough positive evidence for believing or disbelieving. In an evidentialist picture (that supports the view that a person is justified in a doxastic attitude if it fits the given evidence)
one could say that ``suspension of judgment is the justified attitude when the person's evidence on balance supports neither a proposition nor its negation'' \citep[][p.~75]{feldman.2018}. In some cases, we might just have no or barely any evidence for or against a proposition $p$. In other cases, we might have evidence, but the evidence is (almost) equally balanced. For example, we might have evidence for believing $p$: ``There is a Husky in the image,'' because the dog in the image seems white. We might also have evidence for disbelieving $p$, because, the dog seems not to have blue eyes. In such situations, suspension functions as a fallback position that we are justified in when we are not justified in any other doxastic attitude. We are then justified privatively.\\
\indent
We see that epistemologists describe at least two different kinds of reasons for suspension. One type of reason consists of reasons that positively speak for suspension, the other type of reason occurs when one has neither reason to believe nor reason to disbelieve, or in other words: no reason to choose one of the defined answers to a question.\\
\newline
In AML systems we find a correspondence with both these types of justifications. When we look at the justifications for abstention, we can observe that the different justifications considered in epistemology are at play in abstention for ambiguity cases and in abstention for outlier cases. When a system abstains due to ambiguity, a particular input data point is considered ambiguous, meaning that the point is in a region where two (or more) classes overlap. In ambiguity abstention, we have positive evidence for class A \textit{and} positive evidence for class B. For example, some features of the input image speak for the class \texttt{Husky}, while others speak for the class \texttt{Retriever}. Therefore, we abstain in a \textit{privative way}. There is no positive evidence for abstention, but conflicting evidence for different classes.\\
\indent
This is different for outlier abstention. Here, the system abstains from classifying an input sample \emph{because} the sample is an outlier. The sample being an outlier is \textit{positive evidence} for abstention on that sample. This can be seen in cases of pre-attached abstention, where it is decided that abstention is the correct output even before the system is queried about the question and the possible defined answers. This particular input or question is not to be decided by the algorithm.
Hence, we can conclude that both, cases of being neutral due to privative reasons and cases of being neutral due to positive reasons are present in AML systems. In conclusion, the results for the reasons for suspension and abstention are summarized in Table \ref{AML-Table-ReasonsandJustification}.
\begin{table}[H] 
    \centering
    \begin{tabularx}{\textwidth}{XX}
        \hline
    \makecell[l]{Reasons for Abstention\\in Machine Learning}
      & \makecell[l]{Justification for Suspension\\ in Philosophy}  \\
    \hline
Ambiguity Abstention & Privative Justification\\
Outlier Abstention & Positive Justification \\
    \hline
    \end{tabularx}
    \caption{Different reasons for abstention in AML and different corresponding justifications for suspension.}
    \label{AML-Table-ReasonsandJustification}
\end{table}
\subsubsection{Nature of Suspension and Abstention}\label{AML-Suspension-Nature}
The more complex question pertains to the relationship between the nature of suspension and the implementation of abstention in AML systems. In the broader context of assessing the actual ``intelligence'' of various AI systems and their ability to mimic human reasoning processes, it is crucial to explore whether different AML systems can mimic what we call ``suspension of judgment'' when abstaining on a specific question.
To address this question, we must delve into how philosophers characterize the phenomenon of suspension that we experience in human life every day.
\\
\indent
A good way to start investigating these topics is to precisely describe which question is addressed by suspending or abstaining. As described earlier, suspension can be characterized as a way of behaving doxastically to a question under discussion (or an answer to the question) \citep{friedman.2013question, archer.2018, wagner.2022}. This means that suspension is one way of responding to a QUD $Q$ by \textit{not choosing} one of the defined answers. Suspension is characterized as one possible position towards the question under discussion, e.g.,\ ``What kind of dog is on this image?'', different from both belief and disbelief. In the classical picture suspension is one of three doxastic positions in the doxastic triad consisting of belief, disbelief, and suspension.
\\
\newline
Basically, abstention in ML algorithms describes a similar phenomenon, namely the generation of an output with respect to a question that does not match any of the defined answers.\\
\newline
In the case of attached systems, the analogy between suspension and abstention can be drawn only to a limited extent, though.
In attached systems, two different questions play a role in generating the abstention output. One question is the actual question under discussion, i.e.,\ ``Which kind of dog is in the image?'' that is to be answered by the predicting algorithm.  The second question is of the type: ``Is the (possible) answer to the first question certain enough?''\\
\indent
In the case of post-algorithmic attachment, the question under discussion is answered first. This is done in a conventional sense, i.e.,\ in exactly the same way as in a non-abstaining system. A defined answer (e.g.\ \texttt{Husky}) is generated.\footnote{One has to acknowledge that the answer is more informative than just choosing one class, i.e.,\ when the question is answered, there is more information present, e.g., about the probability for this answer being the correct one, and about the probability for other answers.} Only afterward the second question (``Is this answer certain enough?'') is asked. This is the question that is answered by the abstaining part of the algorithm. Hence, in this picture, abstention is not a response or attitude towards the question under discussion, but a response to the second question asked about the certainty of the first answer.
In the case of pre-algorithmic attachment, we find a similar situation, but the order of the questions is reversed.\\
\indent
Therefore, for attached systems, the analogy between suspension and abstention fails in so far as suspension is supposed to address the same questions as the other possible doxastic attitudes. Suspension is a response towards the question under discussion. Abstention in attached systems is an answer to a different question than the question under discussion.\\
\newline
This is different for merged systems. Here, abstention is considered an extra class among the other options for classification. Thus, abstention is one response to the question under discussion. The system is asked: ``What kind of dog is on this image?'' and responds either by providing a defined answer (e.g.\ \texttt{Husky}) as the output class or responds by choosing the abstaining output class. As described earlier, abstention and prediction are parts of the same process and occur simultaneously. Thus, abstention addresses the question under discussion directly. 
\\
\indent
In addition, the different implementations of merged systems (labeled vs. unlabeled) can also be compared with different forms of suspension found in the philosophical literature. 
For example, \citet{ferrari.2022} distinguish between epistemic suspension and indeterminacy suspension (among others).\footnote{\citet{ferrari.2022} take the term agnosticism to refer to the broad concept that subsumes different versions. We take suspension to be this broad term. Hence, we will use the term suspension in the following when \citet{ferrari.2022} would talk about agnosticism.} This distinction consists of different attitudes as to whether the question under discussion is in general answerable or not. One stereotypical case for indeterminacy suspension is a case of mathematical indeterminacy for which a subject can conclude that the proposition is in fact neither true nor false but ontologically indeterminate.\footnote{The most prominent case is the continuum hypothesis \citep{goedel.1947}.}
In cases of epistemic suspension, the subject will take the question in principle to be decidable, but not according to their current epistemic stance.\footnote{It is important to emphasize that in \citet{ferrari.2022}, both epistemic and indeterminacy suspension are regarded as ``pessimistic'' forms of suspension, indicating that the subject does not believe that further inquiry will ultimately resolve the question in a positive or negative manner. Nonetheless, when suspending epistemically, the subject believes that a better evidential situation could, in principle, lead to answering the question, although being pessimistic about reaching that better situation when continuing to inquire.}
\\
\indent
This difference in attitude regarding the question is also found to some degree in the labeled and unlabeled implementations of the merged systems. On the one hand, we have merged systems that learn abstention in a labeled way. We externally tell the system in the training phase which input data (e.g.,\ images) should trigger the response \texttt{abstention}. Here, \texttt{abstention} is considered one ground-truth label of the image. In a certain sense, we ascribe an indeterminate state to these images, which is supposed to be accompanied by abstention. We basically say, no matter how the parameters of the classifier are selected, this image is not to be classified (by a defined answer or label).\\
\indent
Moreover, abstention in such an implementation no longer exactly fulfills the role we ascribed to it in the description of the overarching phenomenon. We described both suspension and abstention as ways of responding to a question \textit{without} selecting one of the defined answers. We diverge from this picture when abstention is learned in a labeled way. Then, abstention no longer represents the non-selection of a defined answer but represents a defined answer itself. In the training phase, abstention is treated analogously to the other classes: the abstention output is \textit{learned} in exactly the same way as the other outputs. The loss calculated for misclassifying a point with the label \texttt{abstention} is conceptually equal to that of misclassifying a point with any other label.
By labeling certain training data as \texttt{abstention}, we treat abstention as a regular class among the others and, thus, as one of the defined answers.\\
\indent
\citet{ferrari.2022} draw a similar picture regarding indeterminacy suspension. They argue that this kind of suspension could be argued to not count as suspension at all if the question is opened to the extent that indeterminacy is one of the conventional, defined answers. The answer set is just expanded, such that it can account for indeterminacy cases. However, choosing this answer is no different from choosing any other answer.\\
\newline
In merged systems, in which abstention is learned in an unlabeled way, the situation is different. Here, abstention is also a possible output class, but it has a special role compared to the other classes. The abstaining response addresses the question in a different way than the other outputs (the defined answers). Abstention is not learned by explicit abstention prototypes, but by giving the system the option not to select any of the other classes in cases of unclear data. In this case, abstention is a way of opting out of choosing one of the defined answers. It reflects epistemic uncertainty. There is uncertainty about the correct defined answer, but it is not assumed that the correct defined answer could not be found in a better evidential situation, or that the correct answer to this question \emph{is} \texttt{abstention}. This is similar to the case of epistemic suspension.\\
\newline
This special role of abstention also aligns well with characterizations of suspension in the philosophical literature. Many authors posit that suspension, as the third doxastic attitude, is more sophisticated and holds a special role compared to belief and disbelief \citep{wedgwood.2002, crawford.2004, Friedman.2013cSuspended, friedman.2017, raleigh.2021, mcgrath.2021, wagner.2022}. According to scholars like \citet{crawford.2004, bergmann.2005, rosenkranz.2007, raleigh.2021, wagner.2022}, the distinctive nature of suspension, in contrast to its doxastic counterparts of belief and disbelief, lies in its status as a \textit{higher-order attitude}. In this view, suspension presupposes indecision, which is then qualified as suspension by the subject either by forming a belief about this uncertainty \citep{crawford.2004, raleigh.2021} or by endorsing the indecision \citep{wagner.2022}. Among others, \citet[][p.~2455]{raleigh.2021} defends a so called \emph{meta-cognitive} view on suspension and asserts that ``suspending whether $p$ constitutively requires having a belief or opinion that one cannot yet tell whether or not $p$, based on one’s evidence'' and that ``such a meta-cognitive opinion about what one can currently tell concerning some question plausibly requires some degree of cognitive sophistication.''
In this perspective, suspension assumes a special role as it necessitates an evaluation of whether one can believe or choose one of the defined answers to a question. This process is more sophisticated and demanding than simply believing one of the answers.\\
\indent
In a parallel manner, abstention in unlabeled merged systems plays a special role compared to all other standard output choices. This is characterized by a certain overview when recognizing that choosing one of the defined answers would be problematic.
The parallel is especially evident during the learning phase of these systems. Although the system is assigned the task of determining a predefined regular answer for all data points, in certain cases, it evaluates that abstaining is a more favorable option (in terms of cost) than providing a specific answer.
\\
\newline
It might be argued that the meta-cognitive form of suspension, which consists of a belief about the own evidential situation, can be found in attached systems, too. (Post-) attached systems can be said to evaluate their evidential situation in terms of probabilities or certainty for specific outputs. While this process might have a meta-cognitivist appearance, it is distinct from what philosophers have in mind when talking about suspension being meta-cognitive. For suspension as a meta-cognitive attitude, there \textit{first} must be indecision as such, which is \textit{then} evaluated by a kind of introspection on a second level.\footnote{The connection between indecision and the second-order belief is different in the account presented by \citet{raleigh.2021}. In his model, the second-order belief is constitutive for indecision and, in this context, takes precedence. Nevertheless, the crucial point is that, in practice, all of these approaches involve a state of indecision concerning the proposition $p$.} For post-attached systems, we find two disanalogies with this picture.
First, in post-attached systems, there is no indecision at all, since an answer has de facto already been selected. As we have argued, the question under discussion is here answered in a non-abstaining way by selecting one of the defined answers; abstention addresses a different question than the question that is under discussion. 
Second, it seems arguable whether there really is an evaluation of one's \textit{own} evidential situation. On the contrary, it could be argued that the predicting and abstaining parts are two systems. In this respect, it is difficult to speak of the abstaining part evaluating \textit{its own} evidential situation. The results for how the different implementations of abstention correspond to suspension are summarized in Table \ref{AML-Table-ImplementationsandSuspension}.
\begin{table}[H] 
    \centering
    \begin{tabularx}{\textwidth}{llX}
        \hline
    \makecell[l]{Implementation\\of Abstention}  & \makecell[l]{Qualification\\for Suspension?} & Form of Suspension \\
    \hline
Attached & \emph{no} & -- \\
Merged & \emph{yes} & 
\parbox[t]{10cm}{Indeterminacy for \emph{Labeled Abstention}\\ Epistemic for \emph{Unlabeled Abstention}}\\
\\
    \hline
    \end{tabularx}
    \caption{Correspondence of the different implementations of abstention in AML with the nature of suspension as well as with different forms of suspension.}
    \label{AML-Table-ImplementationsandSuspension}
\end{table}
\noindent
\subsection{Autonomy of Abstaining}\label{AML-Autonomous}
In this section, we aim to explore the autonomy of abstention in various AML systems. The level of autonomy in the outputs of ML systems is an important topic when philosophically assessing the appropriateness of ascribing intelligence to artificial systems \citep{russell.2002}. Consequently, it becomes imperative to examine the autonomy of AML systems, especially concerning their abstaining output.
\indent
The term ``autonomy'' is discussed controversially in the philosophy of AI and is not easy to define. 
Nevertheless, there are two (connected) desiderata that are emphasized repeatedly and that emerge as commonly accepted criteria in debates around autonomous AI. First, the way from the input to the output is \textit{not} supposed to be completely \textit{hard-coded} by the programmer, and second, some kind of \textit{flexible learning} has to be involved.\\
\indent
\citet[][p.~576]{johnson.2017}, for example, define autonomous AI as ``computational artefacts that are able to achieve a goal without having their course of action fully specified by a human programmer'' and claim that ``learning can play a significant role in seeming to expand the autonomy of computational artefacts'' \citep[][p.~583]{johnson.2017}. \citet{anderson.2011} also stress that autonomy can only be present if the behavior of the system is not micro-managed by humans.
\citet[][p.~42]{russell.2002} claim that ``to the extent that an agent relies on the prior knowledge of its designer rather than on its own precepts and learning processes, we say that the agent lacks autonomy.''
\\
\indent
The two criteria are also emphasized in the discussion on artificial agency which is a concept that is closely related to autonomy \citep{russell.2002}. 
As noted in \citet{eva.2022}, a model of an artificial agent has to make sure that the agent is set up in a way such that it can make its own decisions and is not pre-programmed for all actions and all circumstances.
Also, \citet{muller.2018} emphasize that ``free agents have to be learning agents'' and that the learning history of an agent becomes part of the agent's identity and explains the agent's behavior. These learned but flexible behavior patterns make it possible to attribute actions to the agent itself (see also \citet{briegel.2015}).\\
\indent
Apart from these two necessary criteria for artificial agency and autonomy, \citet{bradshaw.2013} emphasize that it makes sense to speak of autonomous \textit{capabilities} rather than of autonomous \textit{systems} as such since there will always be some activities or capabilities of one system that are autonomous while others may not. We agree with this shift of perspective. In this section, we specifically ask about the autonomy of the \emph{abstaining} capability rather than about the autonomy of the predicting activity or the autonomy of the system itself.\\
\newline
To determine the autonomy of the \emph{abstaining} capability of a systems the two minimal demands for autonomy should be assessed for the abstaining activity \textit{in the same way} as for the predicting activity.
This means that we demand that (a) the way in which a system arrives at the abstaining output should not be completely hard-coded and (b) the connection from the input to the output \texttt{abstention} should be in some way learned by the system.\\
\indent
The system should be able to independently establish a correlation between certain aspects of the inputs and an \texttt{abstention} output.
Not all ML systems belonging to the class of abstaining ML meet this requirement. Attached systems typically consist of an ML system that is trained on the data and that is responsible for predicting, \emph{and} an additional rejection part that is responsible for the abstention task.
Thus, in the attached AML systems, the act of abstention is performed by an algorithm that is separate from the algorithm that performs (in a fairly autonomous ML fashion) the task of prediction. Often, the abstention part of the algorithm is itself a simple, hard-coded piece of the program that is not connected to the machine learning part.\footnote{However, it is possible that the attached abstention part involves some kind of learning. For example, the optimal rejection threshold may also be learned \citep{destefano.2000}. Still, this type of learning does not involve (autonomously) establishing a link between the input data and an abstention output.}
Therefore, the kind of autonomy that is present for the predicting capability in ML systems is not present for the abstaining capability in attached AML systems. We can say that attached AML systems do not abstain \textit{as autonomously} as they predict.
\\
\newline
This is different for merged AML systems. Merged abstention systems autonomously abstain to the same extent that (regular) ML systems make decisions autonomously.
In merged systems, the option of abstention is offered in the training phase, and the system establishes a connection between the features of the input data and an abstention output. Though in different ways, this connection is made both in labeled and unlabeled merged systems.
A merged system can be described as learning to identify situations where a prediction is too risky and thus can be viewed as evaluating its own evidential situation independently of the programmer. In this sense, a merged abstention system can be described as ``knowing when it doesn't know'' \citep{thulasidasan.2019}. Note that we do not claim that merged AML systems abstain autonomously, but rather that in contrast to attached systems, they meet the minimal criterion of autonomous abstaining. The abstaining activity is not hard-coded but learned in some way.
Merged AML systems are \emph{as autonomous} in abstention as they are in prediction. 
\subsection{Explainable Abstaining}\label{AML-XAI}
Beyond the issue of autonomy, explainability is a widely debated topic in the field of (the philosophy of) artificial intelligence, often interconnected with concepts such as interpretability and understanding. This subsection is intended to give a first idea of how investigations about the explainability of AI systems can be extended to abstaining ML systems.\\
\indent
One of the four key principles of explainable AI that are established in \citet[][p.~2]{phillips.2020} is the \textit{Explanation Principle}, which states that ``Systems deliver accompanying evidence or reason(s) for all outputs.'' 
This can be issued, for example, in a procedural way (How did the system reach this output?), in a contrastive way (Why did the system output \emph{this} instead of that answer?), in a recourse way (What do we need to change in the input in order to get another output?). Here, we will focus on local (or instance) explanations, i.e.,\ explaining why a \textit{particular} input sample produces a \textit{particular} output \citep{burkart.2021}.\\
\indent
The explanation principle of \citet{phillips.2020} requires \emph{all} outputs to be accompanied by a reason or explanation. Hence, when considering  AML systems, we must apply this demand not only to the defined answers but also to the abstaining output.\footnote{There are certainly cases where we would intuitively demand an explanation for the defined answers but are fine without an explanation for the abstaining output. Abstaining represents precisely the cautious reaction that does not directly provide us with a decision-making aid in any direction. Therefore, it is sometimes not necessary to ask for an explanation for this option, as long as it is seen as a fallback option that can be used when all other options fail. Still, we would become skeptical if it was used too much.}
In particular, if we want to \textit{learn} something from the abstaining response by improving the input data, examining certain characteristics more closely, or making the training data more diverse, it is useful to know \textit{why} the system reports that it cannot make a decision. Some first approaches to provide explanations for abstaining responses can be found in \citet{Artelt.2022a, artelt.2022b, thulasidasan.2019}.\footnote{On a different note, it is also interesting to evaluate how well the AML classifiers do. An explicit approach to provide metrics for evaluating the results of abstaining classifiers can be found in \citet{Ferri.2004}.}\\
\newline
When we ask for a (local) explanation about the system's abstention on a particular input, we ask about \emph{why} the system abstained on that input or about the \emph{reason} for abstaining on this input. Therefore, the explanation should refer back to the input in some way and point out which parts of the input were responsible for the response (abstaining in this case).
For outlier abstention, this is rather trivial. Abstaining on an outlier can always be explained by referring to the relationship between the training data and the input data point that makes the point an outlier. An explanation is always available and not very informative. The more interesting cases are cases of ambiguity abstention. Thus, we will focus on these in the following.\\
\newline
The distinction between merged and attached systems, which we made in Subsection \ref{AML-attachedandmerged} again becomes relevant for this question about explainability because merged and attached systems allow different options for explanations.\\
\indent
In merged systems, it is (in principle) possible to refer back to the characteristics of the input that are responsible for the abstaining output. If we ask for a reason why the system abstains on a particular input, a merged system can provide such an explanation by pointing to particular features of the input sample just as it can point to the input features that are responsible for, e.g., the output \texttt{Husky} or the output \texttt{Dachshund}. This possibility arises from the fact that merged systems learn to associate certain input characteristics with an abstention. The system thus establishes correlations between characteristics of the input data and an abstention label and can provide the reasons (i.e.,\ some characteristics of the input sample) for abstention. This can serve as a local explanation.\\
\indent
While this seems rather obvious for labeled merged systems, it is interesting to see that this possibility is also available for unlabeled systems. For example, \citet{thulasidasan.2019} use visualization techniques like the one of \citet{selvaraju.2017} to visualize the areas in input images that were relevant for abstaining. 
\citet{thulasidasan.2019} tested their (merged) deep abstaining image classifier (``DAC'') for different abstaining situations. They never labeled the training data with \texttt{abstention}. In a first case, they took 10\% of the training data images and randomized the ground-truth labels. Hence, the ground-truth labels of these images were not correct. There was no regularity in the image-label connection. 
For tracking, they included a ``smudge'' on these images with randomized labels.
In a second experiment, they took all the training images of one class (all monkey images) and randomized the labels while not providing any smudge. In comparison to the first experiment, the noise they created here was ``structured.''
In both experiments, they applied a heat map to the test data, which was supposed to visually highlight the areas of the image that are especially relevant for a certain output. In the first experiment, they found that the system established a correspondence between the smudge and the abstention output. In the heat map, the smudge was highlighted as the part of the image that was decisive for the abstention output. In the second experiment, the typical monkey features were highlighted. This means that the system established a correspondence between either the smudge or typical monkey features and an abstention output, even without being provided with labeled prototypical abstaining cases in the training phase.\footnote{A comparable experiment setup can be found in \citet{Barnes.2021}.
The authors also experiment with corrupting the labels of exactly one (or two) classes. In another experiment, the authors simply corrupt a certain percentage of labels from the training data of all classes. \citet[][p.~3]{Barnes.2021} notice that ``in this case, there is no systematic relationship between the input maps and whether the sample is
corrupted or not. For [these] mixedLabels, we would like the CAN [controlled abstention network] to learn to abstain on the corrupted training samples
by identifying them as those that do not behave like the majority of the training samples.''
It is interesting to see that in this setup there is no intended or pre-specified correlation between input features and the abstaining output. Still, when they test the abstaining system and compare it to the results of a non-abstaining, all-knowing oracle, which serves as an upper bound for accuracy, the results in terms of accuracy (i.e.,\ how many test data points are classified correctly) are nearly ideal.}\\
\indent
This shows how even a merged system that learned abstention not through explicitly \texttt{abstention} labeled training data can still find a connection between certain features of the input space and an abstention output. Thus, one can exploit the full range of local explanations that is available for conventional non-abstaining classifiers. Not only heat maps but any explainable method that is available for regular outputs can be applied to these systems.\\
\newline
For attached systems\footnote{As presented here in the post-algorithmic attachment form for ambiguity abstention. Pre-algorithmic attachment can be neglected as this is mostly possible for outlier abstention.} this is not possible. The system does not find any connection between the characteristics of the input and the abstaining output. It merely learns to connect the characteristics of the input with the conventional outputs. The abstention option, however, is imposed on the system afterward. The attached system abstains when issuing a conventional response is associated with too much uncertainty. So, if we ask for the reason why the system abstains for the specific input $\bm{x}$, the answer (and thus explanation) can only be: ``because the certainty for providing a correct answer is below the threshold.'' Of course, the system can give us information beyond that, such as how far the certainty is from the threshold or the exact probabilities for each answer. If the predicting system itself is explainable, we can possibly even get an answer about which characteristics of the input speak for class A and which for class B and thus concoct an explanation for the abstention ourselves (in the sense of ``the system thinks the head region of the dog looks like a Husky, but the tail looks like a Retriever, hence it abstains''). This could then be seen as an indirect explanation (via the reasons or explanations of the different classes).
However, the system itself cannot provide a straightforward, informative reason for the abstention.
Hence, also in terms of explaining the abstaining output, merged systems surpass attached systems, offering more advanced possibilities for providing explanations.
\section{Conclusion}\label{AML-Conclusion}
This paper was focused on a philosophical analysis of abstaining machine learning (AML) systems. AML systems stand out as the closest approximation to what might be termed ``suspending AI'' in the field of machine learning.
AML systems introduce a novel approach for responding to questions (or tasks like classifying) by refraining from selecting one of the defined answers, essentially opting out. This unique feature enables them to communicate uncertain situations effectively and allows to bring a human in the loop when stakes are too high to allow for decisions that are prone to error.\\
\indent
The objectives of this paper were manifold. Firstly, it aimed to shed light on this type of ML systems that has thus far received limited attention, both within the computer science community and especially in the philosophical community. Secondly, it strove to offer an accessible and informative characterization of these systems. Thirdly, it aimed to explore the various forms and norms of suspension within different AI systems. Lastly, the paper pioneered the first philosophical analysis of abstaining machine learning. The inquiry delved into essential questions in the philosophy of AI, especially concerning autonomy and explainability. AML systems have not yet been considered in these discussions. 
Thereby, this paper provided the first philosophical analysis of abstaining machine learning. 
\\
\newline
We have presented and categorized the different AML systems along two dimensions. We distinguished different reasons to abstain and different ways to abstain. We used these distinctions to evaluate the systems based on philosophical demands. It was shown that the different reasons to abstain in ambiguity and outlier abstention find correspondence in different philosophical norms regarding suspension. We have also examined the technical implementation of AML systems, distinguishing between attached and merged systems. We showed that merged systems generally meet the requirements for suspension that are described in philosophy and that different versions of suspension correspond to different implementations of learned abstention (labeled and unlabeled).
We have shown that in artificial systems there is both a possibility to implement a type of abstention that is structurally similar to the other responses and a possibility to implement abstention with a conceptually more sophisticated special role. 
This is of particular interest from a philosophical perspective since a substantial group of philosophers characterize suspension by its sophisticated, distinctive role and its deviation from belief and disbelief.\\
\indent
We have also shown that merged systems exhibit a higher level of autonomy and that these systems have more room for different opportunities to explain the abstention responses. As a result, this philosophical analysis provides compelling reasons for computer scientists to favor the development of such systems.\\
\newline
However, the findings presented here mark just the initial stage of the philosophical analysis of abstaining machine learning. The two aspects of autonomy and explainability should be further explored, and additional topics, e.g., on consciousness and cognition or understanding of AML systems, warrant investigation. Even in the context of autonomy and explainability, it would be interesting to study the relationship with AML from a different perspective. While this study primarily examined how explainable and autonomous abstaining outputs are, one could also investigate the extent to which the mere capacity to abstain already yields a more autonomous or explainable machine. We are confident that the trust in artificial intelligence is strengthened when these systems acknowledge their uncertainty and effectively communicate it.

\bibliography{Literatur-diss.bib}
\end{document}